\documentclass{article}
\usepackage{iclr2017_conference,times}

\usepackage{url}            
\usepackage{booktabs}       
\usepackage{amsfonts}       
\usepackage{amsmath}        
\usepackage{nicefrac}       
\usepackage{microtype}      
\usepackage{subfig}      
\usepackage{soul} 
\usepackage[font={footnotesize}]{caption}
\usepackage{enumitem}
\usepackage[normalem]{ulem}
\usepackage{rotating}

\usepackage[T1]{fontenc}
\usepackage{floatrow}
\newfloatcommand{capbtabbox}{table}[][\FBwidth]
\setlist{nosep}

\usepackage{color}

\definecolor{darkGreen}{rgb}{0, 0.5, 0.2}
\definecolor{orange}{rgb}{1, 0.75, 0.5}

\definecolor{purple}{rgb}{0.5, 0.2, 0.9}

\definecolor{darkRed}{rgb}{0.4, 0.0, 0.2}

\title{Learning to Navigate\\in Complex Environments}

\author{\small{Piotr Mirowski\thanks{Denotes equal contribution} , \ Razvan Pascanu$^\ast$, \ Fabio Viola, \ Hubert Soyer, \ Andrew J. Ballard,} 
\vspace{0.1 cm}
\\\textbf{\small{Andrea Banino, \ Misha Denil, \ Ross Goroshin, \ Laurent Sifre, \ Koray Kavukcuoglu,}}
\vspace{0.1 cm}
\\\textbf{\small{Dharshan Kumaran, \ Raia Hadsell}} \\
\vspace{0.05 cm}
\\
DeepMind \\
London, UK 
\vspace{0.15 cm}
\\
\scriptsize\texttt{\{piotrmirowski, razp, fviola, soyer, aybd, abanino, mdenil, goroshin, sifre,} \\
\scriptsize\texttt{korayk, dkumaran, raia\} @google.com} \\
}
\begin{document}

\maketitle

\begin{abstract}

Learning to navigate in complex environments with dynamic elements is an important milestone in developing AI agents. 
In this work we formulate the navigation question as a reinforcement learning problem and show that data efficiency and task performance 
can be dramatically improved by relying on additional auxiliary tasks leveraging multimodal sensory inputs. 
 In particular we consider jointly learning the goal-driven reinforcement learning problem with auxiliary depth prediction and loop closure classification tasks. This approach can learn to navigate from raw sensory input in complicated 3D mazes,
approaching human-level performance even under conditions where the goal location changes frequently. We provide detailed
analysis of the agent behaviour\footnote{A video illustrating the navigation agents is available at: \url{https://youtu.be/JL8F82qUG-Q}}, its ability to localise, and its network activity dynamics,  showing that the agent implicitly learns key navigation abilities.
\end{abstract}

\section{Introduction}

The ability to navigate efficiently within an environment is fundamental to intelligent behavior. Whilst conventional robotics methods, such as Simultaneous Localisation and Mapping (SLAM), tackle navigation through an explicit focus on position inference and mapping \citep{Dissanayake2001}, here we follow recent work in deep reinforcement learning \citep{Mnih2015,mnih2016a3c} and propose that navigational abilities could emerge as the by-product of an agent learning a policy that maximizes reward. One advantage of an intrinsic, end-to-end approach is that actions are not divorced from representation, but rather learnt together, thus ensuring that task-relevant features are present in the representation. Learning to navigate from reinforcement learning in partially observable environments, however, poses several challenges. 

First, rewards are often sparsely distributed in the environment, where there may be only one goal location. Second, environments often comprise dynamic elements, requiring the agent to use memory at different timescales: rapid one-shot memory for the goal location, together with short term memory subserving temporal integration of velocity signals and visual observations, and longer term memory for constant aspects of the environment (e.g. boundaries, cues).

To improve statistical efficiency we bootstrap the reinforcement learning procedure by augmenting our loss with auxiliary tasks that provide denser training signals that support navigation-relevant representation learning. We consider two additional losses: the first one involves reconstruction of a low-dimensional depth map at each time step by predicting one input modality (the depth channel) from others (the colour channels). This auxiliary task concerns the 3D geometry of the environment, and is aimed to encourage the learning of representations that aid obstacle avoidance and short-term trajectory planning. The second task directly invokes loop closure from SLAM: the agent is trained to predict if the current location has been previously visited within a local trajectory. 

\begin{figure}
    \centering
    \includegraphics[width=0.28\textwidth,bb=0 0 642 478]{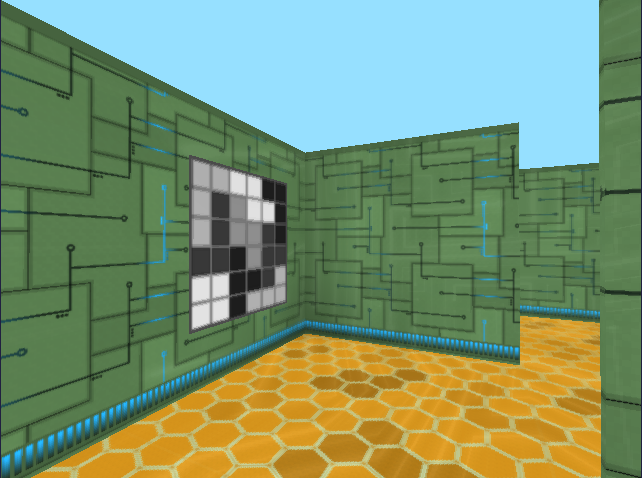}
    \hspace{.05in}
    \includegraphics[width=0.28\textwidth,bb=0 0 639 480]{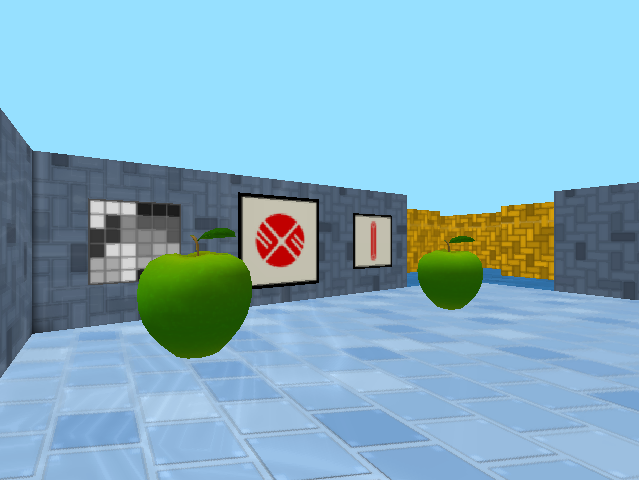}
    \hspace{.05in}
    \includegraphics[width=0.28\textwidth,bb=0 0 641 481]{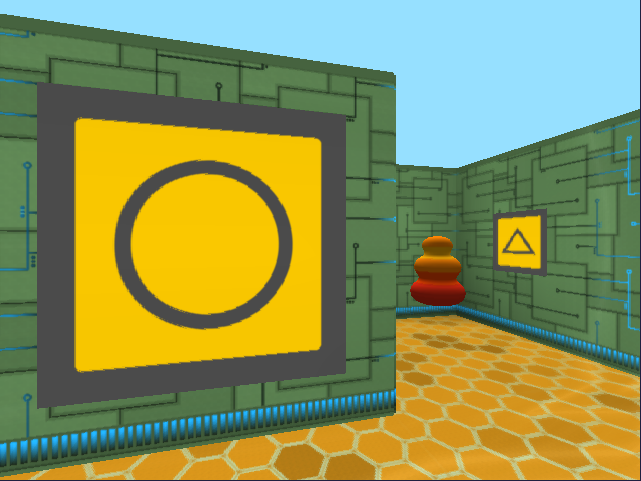}
    \\
    \vspace{.03in}
    \includegraphics[width=0.28\textwidth,height=1in,bb=0 0 1000 500]{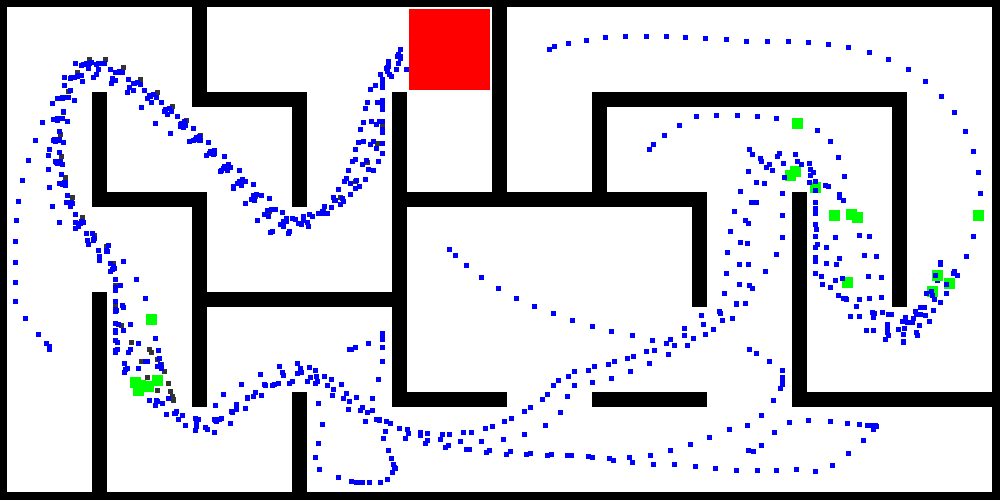}
    \hspace{.05in}
    \includegraphics[width=0.28\textwidth,height=1in,bb=0 0 1500 900]{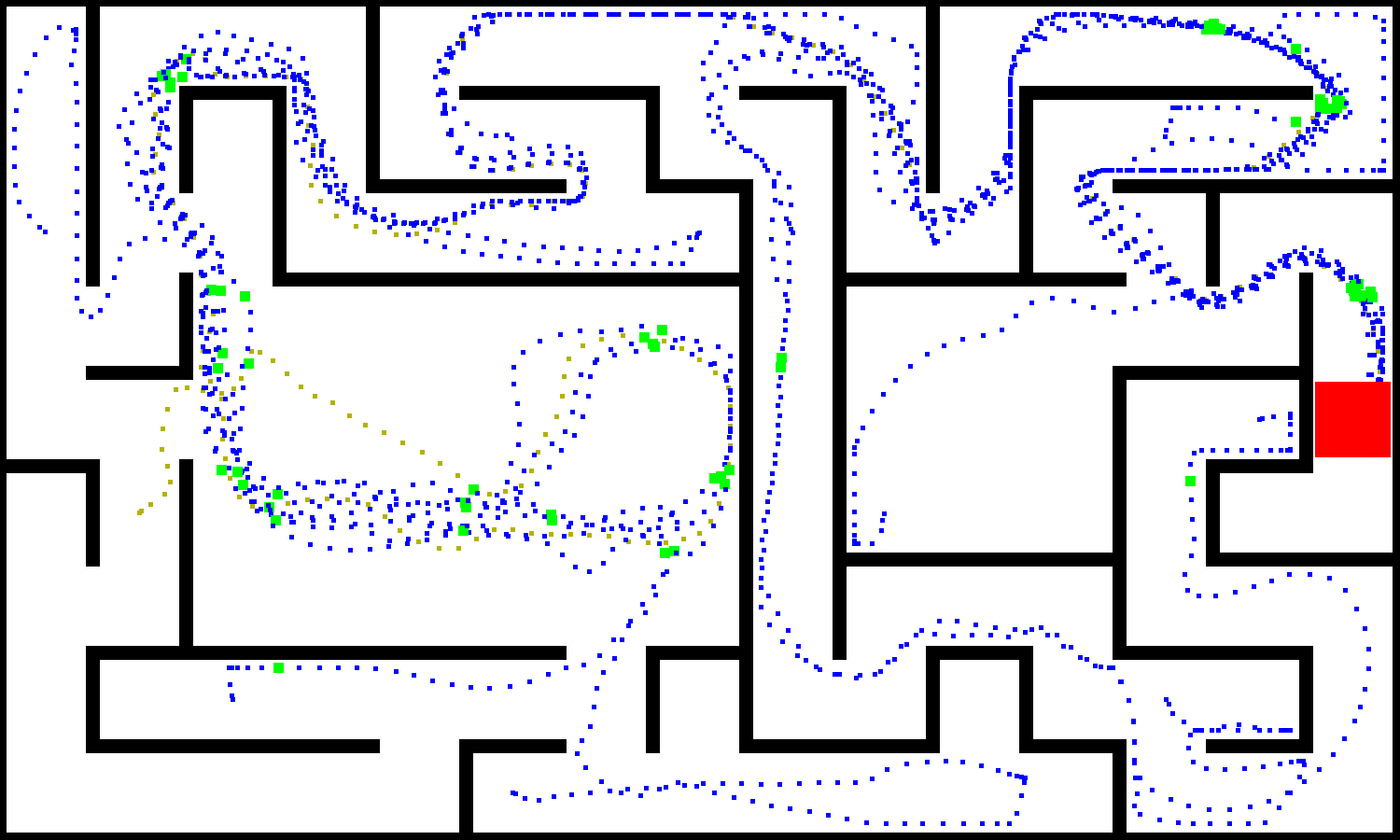}
    \hspace{.05in}
    \includegraphics[width=0.28\textwidth,height=1in,bb=0 0 1100 700]{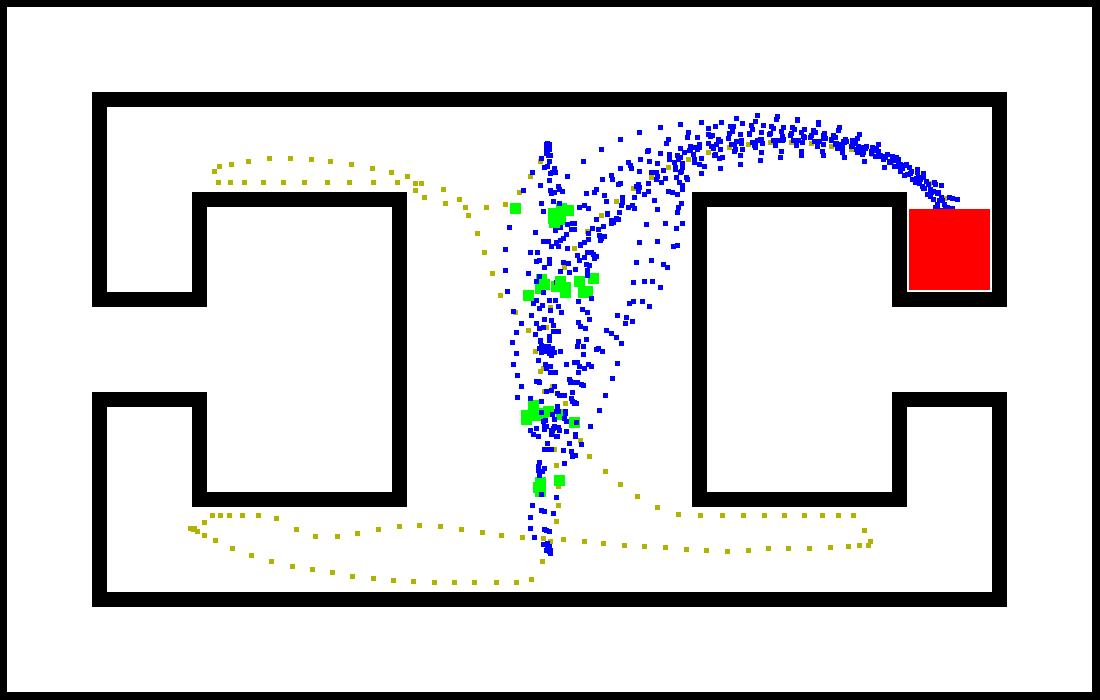}
    \caption{Views from a small $5 \times 10$ maze, a large $9 \times 15$ maze and an I-maze, with corresponding maze layouts and sample agent trajectories. The mazes, which will be made public, have different textures and visual cues as well as exploration rewards and goals (shown right).}
    \label{fig:maze}
\end{figure}

To address the memory requirements of the task we rely on a stacked LSTM architecture \citep{Graves2013,pascanu2013construct}. 
%
We evaluate our approach using five 3D maze environments and demonstrate the accelerated learning and increased performance of the proposed agent architecture. These environments feature complex geometry, random start position and orientation, dynamic goal locations, and long episodes that require thousands of agent steps (see Figure \ref{fig:maze}). We also provide detailed analysis of the trained agent to show that critical navigation skills are acquired. This is important as neither position inference nor mapping are directly part of the loss; therefore, raw performance on the goal finding task is not necessarily a good indication that these skills are acquired. In particular, we show that the proposed agent resolves ambiguous observations and quickly localizes itself in a complex maze, and that this localization capability is correlated with higher task reward. 

\section{Approach}
\label{sec:approach}

We rely on a end-to-end learning framework that incorporates multiple objectives. Firstly it tries to maximize cumulative reward using an actor-critic approach. Secondly it minimizes an auxiliary loss of inferring the depth map from the RGB observation. Finally, the agent is trained to detect loop closures as an additional auxiliary task that encourages implicit velocity integration. 

The reinforcement learning problem is addressed with the Asynchronous Advantage Actor-Critic (A3C) algorithm \citep{mnih2016a3c} that relies on learning both a policy $\pi(a_t|s_t;\theta)$ and value function $V(s_t;\theta_V)$ given 
a state observation $s_t$. Both the policy and value function share all intermediate representations, both being computed 
using a separate linear layer from the topmost layer of the model. 
%
The agent setup closely follows the work of \citep{mnih2016a3c} and we refer to this work for the details (e.g. the use of a convolutional encoder followed by either an MLP or an LSTM, the use of action repetition, entropy regularization to prevent the policy saturation, etc.). These details can also be found in the Appendix~\ref{sec:app}. 

\begin{figure}[ht]
  \centering
    \includegraphics[width=.8\textwidth]{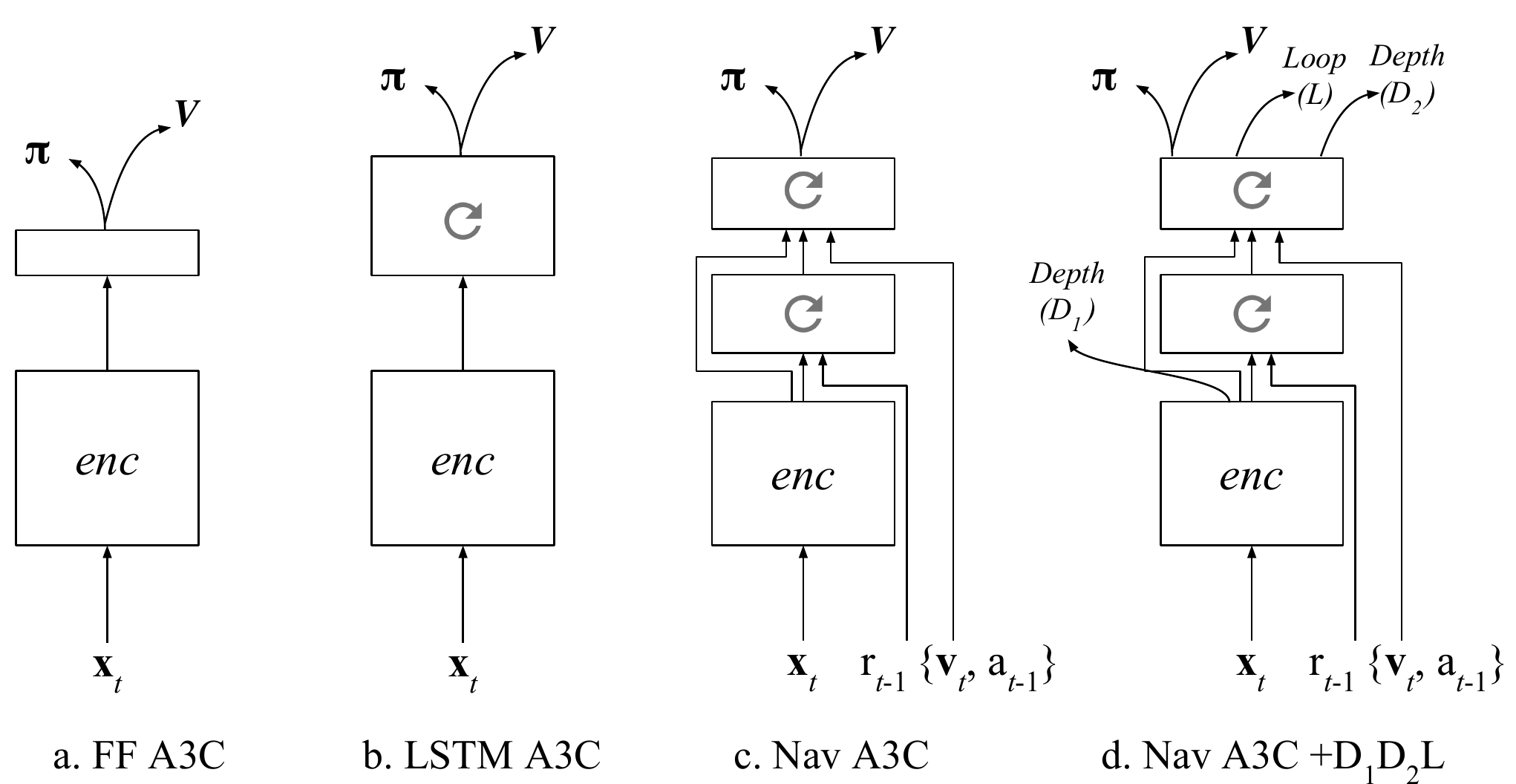}
    \caption{ Different architectures: (a) is a convolutional encoder followed by a feedforward layer and policy ($\pi$) and value function outputs; (b) has an LSTM layer; (c) uses additional inputs (agent-relative velocity, reward, and action), as well as a stacked LSTM; and (d) has additional outputs to predict depth and loop closures.
    }
    \label{fig:archs}
\end{figure}

The baseline that we consider in this work is an A3C agent \citep{mnih2016a3c} that receives only RGB input from the environment, using either a recurrent or a purely feed-forward model (see Figure~\ref{fig:archs}a,b).
The encoder for the RGB input (used in all other considered architectures) is a 3 layer convolutional network. 
To support the navigation capability of our approach, we also rely on the Nav A3C agent (Figure~\ref{fig:archs}c) which employs a two-layer stacked LSTM after the convolutional encoder. We expand the observations of the agents to include \emph{agent-relative} velocity, the action sampled from the stochastic policy and the immediate reward, from the previous time step. We opt to feed the velocity and previously selected action directly to the second recurrent layer, with the first layer only receiving the reward. We postulate that the first layer might be able to make associations between reward and visual observations that are provided as context to the second layer from which the policy is computed. 
Thus, the observation $s_t$ may include an image ${\bf x}_t \in \mathbb{R}^{3\times W \times H}$ (where $W$ and $H$ are the width and height of the image), the agent-relative lateral and rotational velocity  ${\bf v}_t \in \mathbb{R}^{6}$, the previous action ${\bf a}_{t-1} \in \mathbb{R}^{N_A}$, and the previous reward $r_{t-1} \in \mathbb{R}$.

Figure~\ref{fig:archs}d shows the augmentation of the Nav A3C with the different possible auxiliary losses. 
In particular we consider predicting depth from the convolutional layer (we will refer to this choice as $D_1$), or from the top LSTM layer ($D_2$) or predicting loop closure ($L$). The auxiliary 
losses are computed on the current frame via a single layer MLP. The agent is trained  by applying 
a weighted sum of the gradients coming from A3C, the gradients from depth prediction (multiplied with $\beta_{d_1}, \beta_{d_2}$) and the gradients from the loop closure (scaled by $\beta_l$). 
More details of the online learning algorithm are given in Appendix \ref{sec:app}.


\subsection{Depth prediction}

The primary input to the agent is in the form of RGB images. However, depth information, covering the central field of view of the agent, might supply valuable information about the 3D structure of the environment. 
While depth could be directly used as an input, we argue that if presented as an additional loss it is actually more valuable to the learning process.
In particular if the prediction loss shares representation with the policy, it could help build useful features for RL much faster, 
bootstrapping learning. Since we know from \citep{Eigen2014} that a single frame can be enough to predict depth, we know this auxiliary task can be learnt.
A comparison between having depth as input versus as an additional loss is given in Appendix~\ref{sec:additionalResults}, which shows significant gain for depth as a loss. 

Since the role of the auxiliary loss is just to build up the representation of the model, we do not necessarily care about the specific performance obtained or nature of the prediction. We do care about the data efficiency aspect of the problem and also computational complexity. If the loss is to be useful for the main task, we should converge faster on it compared to solving the RL problem (using less data samples), and the additional computational cost should be minimal. To achieve this we use a low resolution  variant of the depth map, reducing the screen resolution to 4x16 pixels\footnote{The image is cropped before being subsampled to lessen the floor and ceiling which have little relevant depth information.}. 

We explore two different variants for the loss. The first choice is to phrase it as a regression task, the most natural choice. While this formulation, combined with a higher depth resolution, extracts the most information, mean square error imposes a unimodal distribution \citep{aaron-pixelrnn-2016}. To address this possible issue, we also 
consider a classification loss, where depth at each position is discretised into 8 different bands.
The bands are non-uniformally distributed such that we pay more attention to far-away objects (details in Appendix~\ref{sec:app}). The motivation for the classification formulation is that while it greatly reduces the resolution of depth, it is more flexible from a learning perspective and can result in faster convergence (hence faster bootstrapping).



\subsection{Loop closure prediction}

Loop closure, like depth, is valuable for a navigating agent, since 
can be used for efficient exploration and spatial reasoning. 
To produce the training targets, we detect loop closures based on the similarity of local position information during an episode, which is obtained by integrating 2D velocity over time. 
Specifically, in a trajectory noted $\{ p_0, p_1, \dots, p_T \}$, where $p_t$ is the position of the agent at time $t$, we define a loop closure label $l_t$ that is equal to 1 if the position $p_t$ of the agent is close to the position $p_{t'}$ at an earlier time $t'$. 
In order to avoid trivial loop closures on consecutive points of the trajectory, we add an extra condition on an intermediary position $p_{t''}$ being far from $p_t$. Thresholds $\eta_1$ and $\eta_2$ provide these two limits. 
Learning to predict the binary loop label is done by minimizing the Bernoulli loss $\mathcal{L}_l$ between $l_t$ and the output of a single-layer output from the hidden representation $h_t$ of the last hidden layer of the model, followed by a sigmoid activation. 

\section{Related work}

There is a rich literature on navigation, primarily in the robotics literature. However, here we focus on related work in deep RL. Deep Q-networks (DQN) have had breakthroughs in extremely challenging domains such as Atari \citep{Mnih2015}. 
Recent work has developed on-policy RL methods such as advantage actor-critic that use asynchronous training of multiple agents in parallel \citep{mnih2016a3c}. Recurrent networks have also been successfully incorporated to enable state disambiguation in partially observable environments \citep{Gomez2013,HausknechtS15,mnih2016a3c,NarasimhanKB15}. 

Deep RL has recently been used in the navigation domain.  \cite{KulkarniSuccessor16} used a feedforward architecture to learn deep successor representations that enabled behavioral flexibility to reward changes in the MazeBase gridworld, and provided a means to detect bottlenecks in 3D VizDoom. \cite{Zhu16} used a feedforward siamese actor-critic architecture incorporating a pretrained ResNet to support navigation to a target in a discretised 3D environment. \cite{honglakleeICML16} investigated the performance of a variety of networks with external memory \citep{weston2014memory} on simple navigation tasks in the Minecraft 3D block world environment. \cite{Tessler16} also used the Minecraft domain to show the benefit of combining feedforward deep-Q networks with the learning of resuable skill modules (cf options: \citep{sutton1999between}) to transfer between navigation tasks. \cite{Tai2016} trained a convnet DQN-based agent using depth channel inputs for obstacle avoidance in 3D environments. \cite{Barron2016} investigated how well a convnet can predict the depth channel from RGB in the Minecraft environment, but did not use depth for training the agent.

Auxiliary tasks have often been used to facilitate representation learning \citep{suddarth1990rule}. Recently, the incorporation of additional objectives, designed to augment representation learning through auxiliary reconstructive decoding pathways \citep{ZhangICML16,RasmusBHVR15,ZhaoMGL15,Mirowski2010}, has yielded benefits in large scale classification tasks. In deep RL settings, however, only two previous papers have examined the benefit of auxiliary tasks. Specifically, \cite{Li2016} consider a supervised loss for fitting a recurrent model on the hidden representations to predict the next observed state, in the context of imitation learning of sequences provided by experts, and \cite{LampleC16} show that the performance of a DQN agent in a first-person shooter game in the VizDoom environment can be substantially enhanced by the addition of a supervised auxiliary task, whereby the convolutional network was trained on an enemy-detection task, with information about the presence of enemies, weapons, etc., provided by the game engine. 

In contrast, our contribution addresses fundamental questions of how to learn an intrinsic representation of space, geometry, and movement while simultaneously maximising rewards through reinforcement learning. Our method is validated in challenging maze domains with random start and goal locations.

\section{Experiments}



We consider a set of first-person 3D mazes from the DeepMind Lab environment \citep{Beattie2016} (see Fig. \ref{fig:maze}) that are visually rich, with additional observations available to the agent such as inertial information and local depth information.\footnote{The environments used in this paper are publicly available at \url{https://github.com/deepmind/lab}.} The action space is discrete, yet allows finegrained control, comprising 8 actions: the agent can rotate in small increments, accelerate forward or backward or sideways, or induce rotational acceleration while moving. Reward is achieved in these environments by reaching a goal from a random start location and orientation. If the goal is reached, the agent is respawned to a new start location and must return to the goal. The episode terminates when a fixed amount of time expires, affording the agent enough time to find the goal several times. 
There are sparse `fruit' rewards which serve to encourage exploration. Apples are worth 1 point, strawberries 2 points and goals are 10 points. Videos of the agent solving the maze are linked in Appendix \ref{sec:videos}.

In the static variant of the maze, the goal and fruit locations are fixed and only the agent's start location changes. In the dynamic (Random Goal) variant, the goal and fruits are randomly placed on every episode. Within an episode, the goal and apple locations stay fixed until the episode ends. This encourages an explore-exploit strategy, where the agent should initially explore the maze, then retain the goal location and quickly refind it after each respawn. For both variants (static and random goal) we consider a small and large map. The small mazes are $5 \times 10$ and episodes last for 3600 timesteps, and the large mazes are $9 \times 15$ with 10800 steps (see Figure \ref{fig:maze}). The RGB observation is $84 \times 84$.

The I-Maze environment (see Figure \ref{fig:maze}, right) is inspired by the classic T-maze used to investigate navigation in rodents \citep{olton1979hippocampus}: the layout remains fixed throughout, the agent spawns in the central corridor where there are apple rewards and has to locate the goal which is placed in the alcove of one of the four arms. Because the goal is hidden in the alcove, the optimal agent behaviour must rely on memory of the goal location in order to return to the goal using the most direct route. Goal location is constant within an episode but varies randomly across episodes.


\begin{figure}
    \centering
    \subfloat[\scriptsize{Static maze (small)}]{
    \includegraphics[width=0.33\textwidth,bb=0 0 505 355]{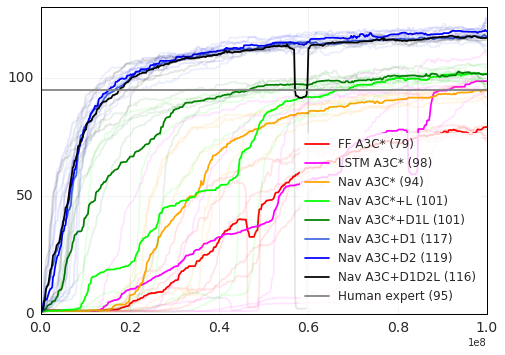} 
    }
    \subfloat[\scriptsize{Static maze (large)}]{
    \includegraphics[width=0.33\textwidth,bb=0 0 494 355]{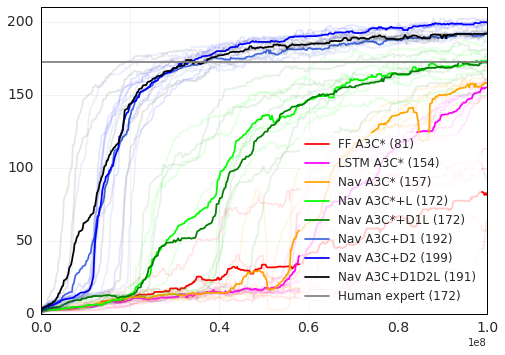} 
    }
    \subfloat[\scriptsize{Random Goal I-maze}]{
    \includegraphics[width=0.33\textwidth,bb=0 0 505 355]{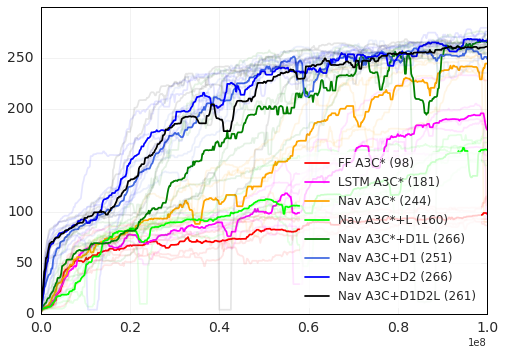} 
    }\\
    \subfloat[\scriptsize{Random Goal maze (small)}]{
    \includegraphics[width=0.33\textwidth,bb=0 0 496 355]{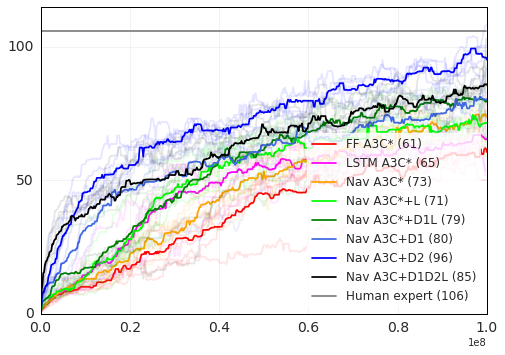} 
    }
    \subfloat[\scriptsize{Random Goal maze (large)}]{
    \includegraphics[width=0.33\textwidth,bb=0 0 505 355]{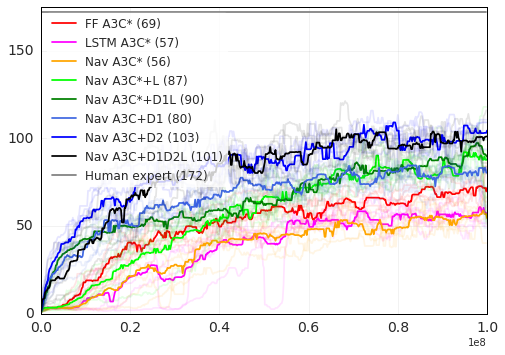} 
    }
    \subfloat[\scriptsize{Random Goal maze (large): different formulation of depth prediction}]{
    \includegraphics[width=0.33\textwidth,bb=0 0 505 355]{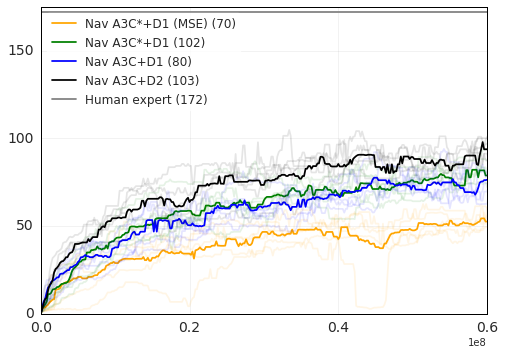}
    }
    \caption{Rewards achieved by the agents  on 5 different tasks: two static mazes (small and large) with fixed goals, two static mazes with comparable layout but with dynamic goals and the I-maze. Results are averaged over the top 5 random hyperparameters for each agent-task configuration. Star in the label indicates the use of reward clipping. Please see text for more details.}
    \label{fig:rewards}
\end{figure}

The different agent architectures described in Section \ref{sec:approach} are evaluated by training on the five mazes. Figure \ref{fig:rewards} shows learning curves (averaged over the 5 top performing agents). The agents are a feedforward model (FF A3C), a recurrent model (LSTM A3C), the stacked LSTM version with velocity, previous action and reward as input (Nav A3C), and Nav A3C  with depth prediction from the convolution layer (Nav A3C+$D_1$), Nav A3C with depth prediction from the 
last LSTM layer  (Nav A3C+$D_2$), Nav A3C with loop closure prediction (Nav A3C+$L$) as well as the Nav A3C  with all auxiliary losses considered together (Nav A3C+$D_1D_2L$).
In each case we ran 64 experiments with randomly sampled hyper-parameters (for ranges and details please see the appendix). The mean over the top 5 runs as well as the top 5 curves are plotted. Expert human scores, established by a professional game player, are compared to these results. The Nav A3C+$D_2$ agents reach human-level performance on Static 1 and 2, and attain about 91\% and 59\% of human scores on Random Goal 1 and 2.

In \cite{Mnih2015} reward clipping is used to stabilize learning, technique which we employed in this work as well.
Unfortunately, for these particular tasks, 
this yields slightly suboptimal policies because the agent does not distinguish 
 apples (1 point) from goals (10 points). 
 Removing 
the reward clipping results in unstable behaviour for the base A3C agent (see Appendix~\ref{sec:additionalResults}). However it seems that the auxiliary signal from depth prediction 
mediates this problem to some extent, resulting in stable learning dynamics  (e.g. Figure~\ref{fig:rewards}f, Nav A3C+$D_1$ vs Nav A3C*+$D_1$). 
We clearly indicate whether reward clipping is used by adding an asterisk to the agent name. 

Figure~\ref{fig:rewards}f also explores the difference between the two formulations of depth prediction, as a regression task or a classification task. We can see that the regression agent (Nav A3C*+$D_1$[MSE]) performs worse than one that does classification (Nav A3C*+$D_1$). This result extends to other maps, and we therefore only use the classification formulation in all our other results\footnote{An exception is the Nav A3C*+$D_1L$ agent on the I-maze (Figure~\ref{fig:rewards}c), which uses depth regression and reward clipping. While it does worse, we include it because some analysis is based on this agent.}. Also we see that predicting depth from the last LSTM layer (hence providing structure to the recurrent layer, not just the convolutional ones) performs better.


We note some particular results from these learning curves. In Figure \ref{fig:rewards} (a and b), consider the feedforward A3C model (red curve) versus the LSTM version (pink curve). Even though navigation seems to intrinsically require memory, as single observations could often be ambiguous, the feed-forward model achieves competitive performance on static mazes. This suggest that there might be good strategies that do not involve temporal memory and give good results, namely a reactive policy held by the weights of the encoder, or learning a wall-following strategy. This motivates the dynamic environments that encourage the use of memory and more general navigation strategies. 

Figure \ref{fig:rewards} also shows the advantage of adding velocity, reward and action as an input, as well as the impact of using a two layer LSTM (orange curve vs red and pink). Though this agent (Nav A3C) is better than the simple architectures, it is still relatively slow to train on all of the mazes. We believe that this is mainly due to the slower, data inefficient learning that is generally seen in pure RL approaches. Supporting this we see that adding the auxiliary prediction targets of depth and loop closure (Nav A3C+$D_1D_2L$, black curve) speeds up learning dramatically on most of the mazes (see Table 1: AUC metric). It has the strongest effect on the static mazes because of the accelerated learning, but also gives a substantial and lasting performance increase on the random goal mazes.

\begin{figure}
    \centering
    \includegraphics[height=1in,bb=0 0 33 50]{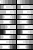}
    \hspace{.1in}
    \includegraphics[height=1in,bb=0 0 33 50]{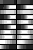}
    \hspace{.3in}
    \includegraphics[height=1in,bb=0 0 1500 900]{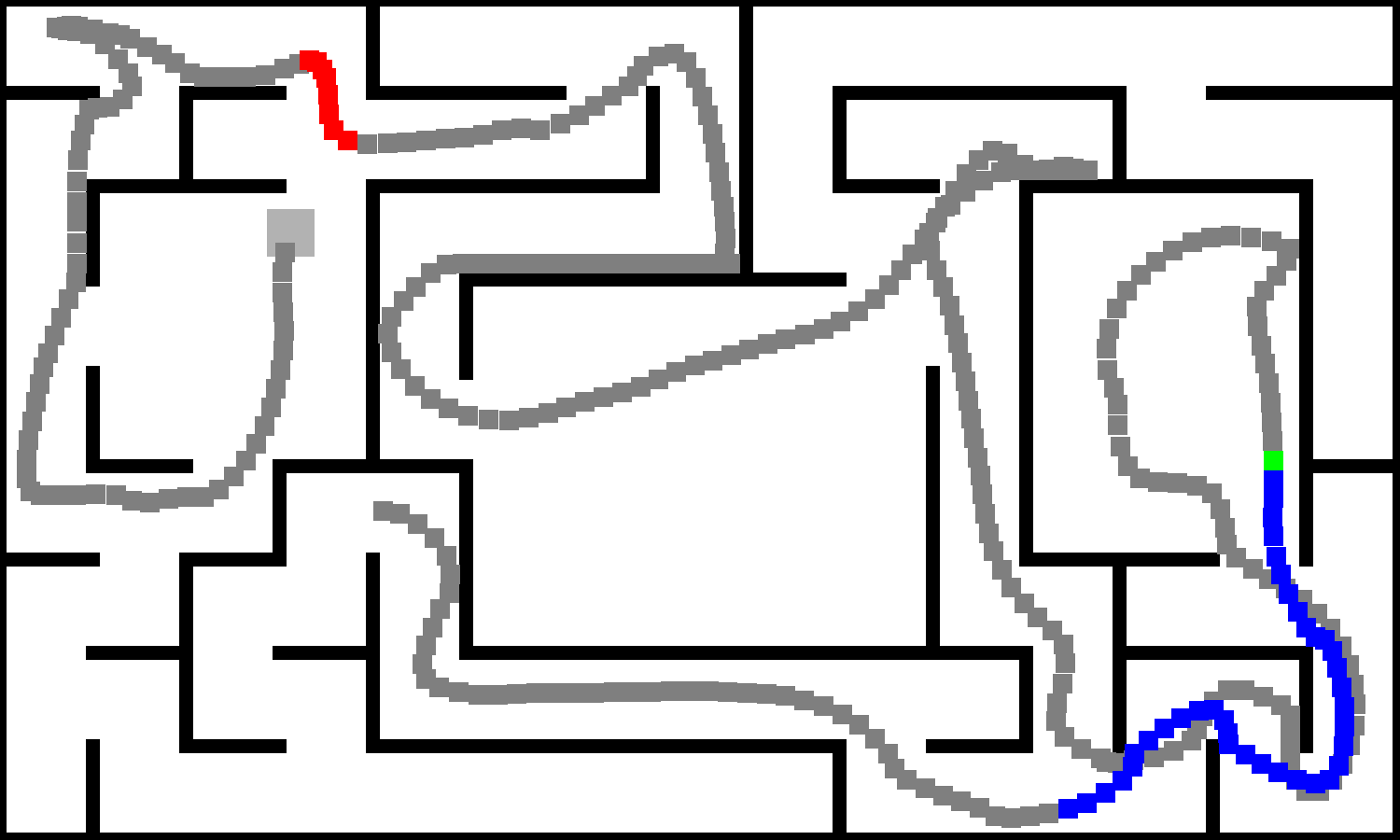}
    \caption{\emph{left}: Example of depth predictions (pairs of ground truth and predicted depths), sampled every 40 steps. \emph{right:} Example of loop closure prediction. The agent starts at the gray square and the trajectory is plotted in gray. Blue dots correspond to true positive outputs of the loop closure detector; red cross correspond to false positives and green cross to false negatives. Note the false positives that occur when the agent is actually a few squares away from actual loop closure.}
    \label{fig:loops}
\end{figure}

Although we place more value on the task performance than on the auxiliary losses, we report the results from the loop closure prediction task. Over 100 test episodes of 2250 steps each, within a large maze (random goal 2), the Nav A3C*+$D_1L$ agent demonstrated very successful loop detection, reaching an F-1 score of 0.83. A sample trajectory can be seen in Figure \ref{fig:loops} (right).


\section{Analysis}

\subsection{Position decoding}
\label{ref:pos}

\begin{table}
\begin{center}
\scriptsize
 \begin{tabular}{@{}ll|ccc|cccc@{}} 
 \toprule
 & & \multicolumn{3}{c|}{Mean over top 5 agents} & \multicolumn{4}{c}{Highest reward agent} \\
 \textbf{Maze} & \textbf{Agent} & \textbf{AUC} & \textbf{Score} & \textbf{\% Human} & \textbf{Goals} & \textbf{Position Acc} & \textbf{Latency 1:>1} & \textbf{Score} \\ 
 \midrule
 \textbf{I-Maze} & FF A3C* & 75.5 & 98 & - & 94/100 & 42.2 & 9.3s:9.0s & 102 \\ 
& LSTM A3C* & 112.4 & 244 & - & {\bf 100/100} & {\bf 87.8} & 15.3s:3.2s & 203 \\
& Nav A3C*+$D_1L$ & 169.7 & 266 & - & {\bf 100/100} & 68.5 & 10.7s:2.7s & 252 \\
& Nav A3C+$D_2$ & {\bf 203.5} & {\bf 268} & - & {\bf 100/100} & 62.3 & {\bf 8.8s:2.5s} & {\bf 269} \\
& Nav A3C+$D_1D_2L$ & 199.9 & 258 & - & {\bf 100/100} & 61.0 & {\bf 9.9s:2.5s} & 251 \\
 \midrule
 \textbf{Static 1} & FF A3C* & 41.3 & 79 & 83  & 100/100 & 64.3 & 8.8s:8.7s & 84 \\ 
  & LSTM A3C* & 44.3 & 98 &  103  & 100/100 & 88.6 & 6.1s:5.9s & 110 \\
  & Nav A3C+$D_2$ & {\bf 104.3} & {\bf 119} & {\bf 125} & 100/100 & {\bf 95.4} & {\bf 5.9s:5.4s} & 122 \\
  & Nav A3C+$D_1D_2L$ & 102.3 & 116 & 122 & 100/100 & 94.5 & {\bf 5.9s:5.4s} & {\bf 123} \\
 \midrule
\textbf{Static 2} & FF A3C* & 35.8 & 81 & 47 & 100/100 & 55.6 & 24.2s:22.9s & 111 \\ 
  & LSTM A3C* & 46.0 & 153 & 91 & 100/100 & 80.4 & 15.5s:14.9s & 155 \\ 
  & Nav A3C+$D_2$ & {\bf 157.6} & {\bf 200} & {\bf 116} & 100/100 & {\bf 94.0} &  10.9s:11.0s & {\bf 202}  \\ 
  & Nav A3C+$D_1D_2L$ & 156.1 & 192 & 112 & 100/100 & 92.6 & 11.1s:12.0s & 192  \\ 
 \midrule
 \textbf{Random Goal 1} & FF A3C* & 37.5 & 61 & 57.5 & 88/100 & 51.8 & 11.0:9.9s & 64 \\ 
  & LSTM A3C* & 46.6 & 65 & 61.3 & 85/100 & 51.1 & 11.1s:9.2s & 66 \\
  & Nav A3C+$D_2$ & {\bf 71.1} & {\bf 96} & {\bf 91} & {\bf 100/100} & {\bf 85.5} & {\bf 14.0s:7.1s} & {\bf 91} \\ 
  & Nav A3C+$D_1D_2L$ & 64.2 & 81 & 76 & 81/100 & 83.7 & 11.5s:7.2s & 74.6 \\  \midrule
  \textbf{Random Goal 2} & FF A3C* & 50.0 & 69 & 40.1 & {\bf 93/100} & 30.0 & 27.3s:28.2s & 77 \\ 
  & LSTM A3C* & 37.5 & 57 & 32.6 & 74/100 & 33.4 & 21.5s:29.7s & 51.3 \\ 
  & Nav A3C*+$D_1L$ & 62.5 & 90 & 52.3 & 90/100 & 51.0 & 17.9s:18.4s & 106 \\ 
  & Nav A3C+$D_2$ & {\bf 82.1} & {\bf 103} & {\bf 59} & 79/100 & 72.4 & {\bf 15.4s:15.0s} & {\bf 109} \\ 
  & Nav A3C+$D_1D_2L$ & 78.5 & 91 & 53 & 74/100 & {\bf 81.5} & 15.9s:16.0s & 102 \\ 
 \bottomrule
\end{tabular}
\end{center}
\caption{Comparison of four agent architectures over five maze configurations, including random and static goals. \emph{AUC} (Area under learning curve), \emph{Score}, and \emph{\% Human} are averaged over the best 5 hyperparameters. Evaluation of a single best performing agent is done through analysis on 100 test episodes. \emph{Goals} gives the number of episodes where the goal was reached one more more times. \emph{Position Accuracy} is the classification accuracy of the position decoder. \emph{Latency 1:>1} is the average time to the first goal acquisition vs. the average time to all subsequent goal acquisitions. \emph{Score} is the mean score over the 100 test episodes.}
\label{tab:position}
\end{table}

In order to evaluate the internal representation of location within the agent (either in the hidden units $h_t$ of the last LSTM, or, in the case of the FF A3C agent, in the features $f_t$ on the last layer of the conv-net), we train a position decoder that takes that representation as input, consisting of a linear classifier with multinomial probability distribution over the discretized maze locations. Small mazes ($5 \times 10$) have 50 locations, large mazes ($9 \times 15$) have 135 locations, and the I-maze has 77 locations. 
Note that we do not backpropagate the 
gradients from the position decoder through the 
rest of the network. The position decoder can 
only see the representation exposed by the model, not change it.

\begin{figure}
    \centering
    \includegraphics[width=0.22\textwidth,bb=0 0 1000 500]{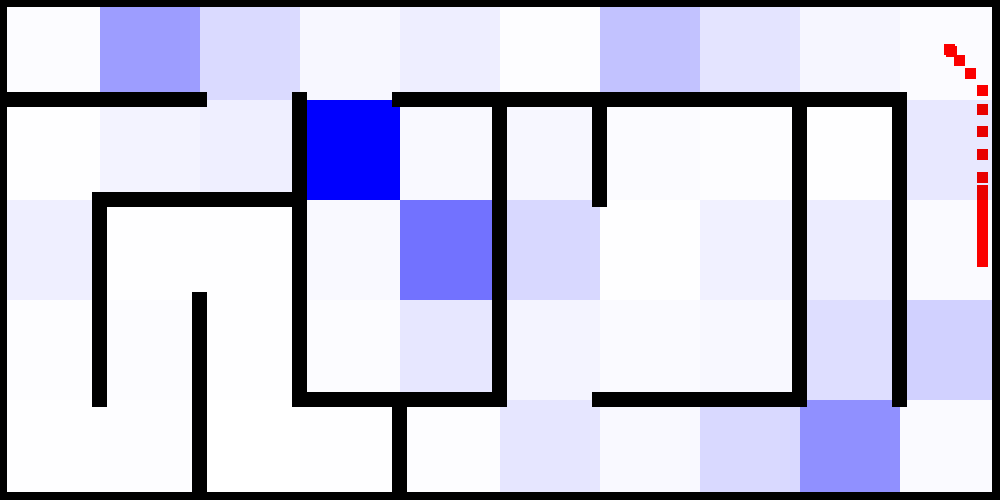}
    \includegraphics[width=0.22\textwidth,bb=0 0 1000 500]{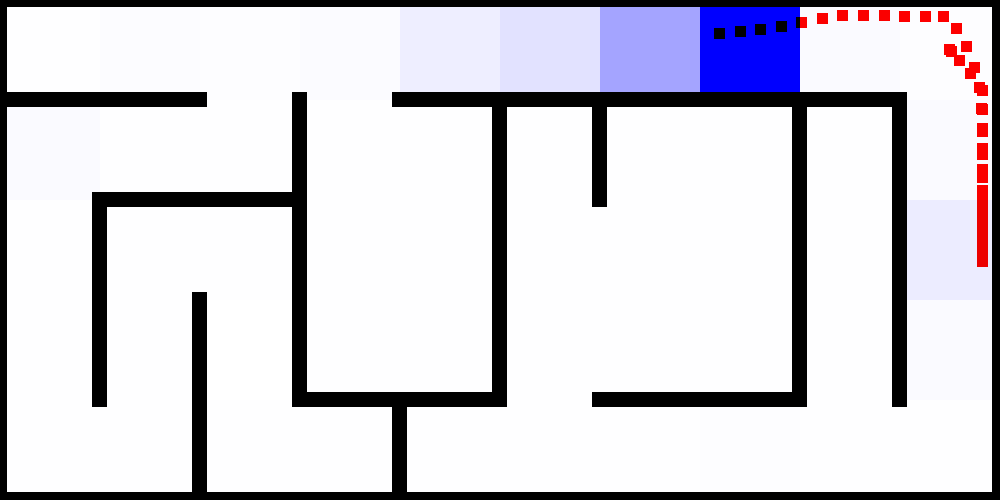}
    \includegraphics[width=0.22\textwidth,bb=0 0 1000 500]{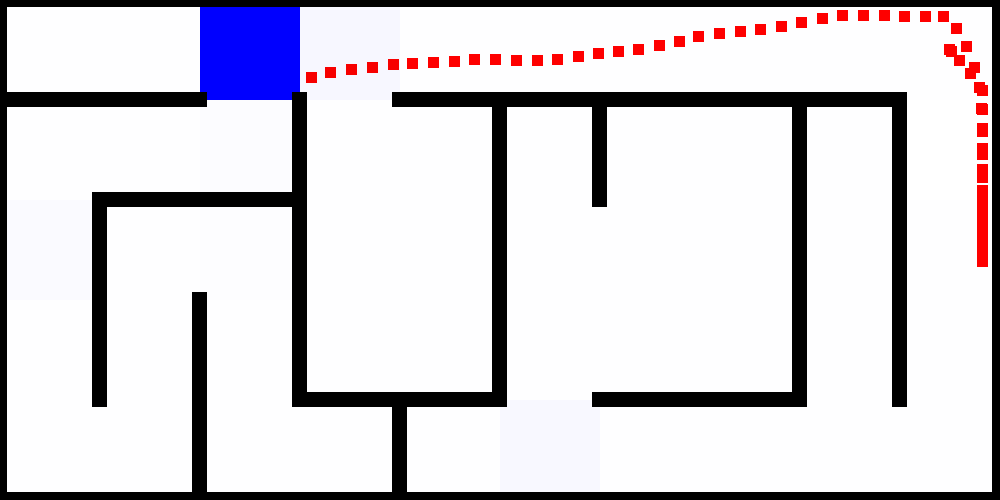}
    \includegraphics[width=0.22\textwidth,bb=0 0 1000 500]{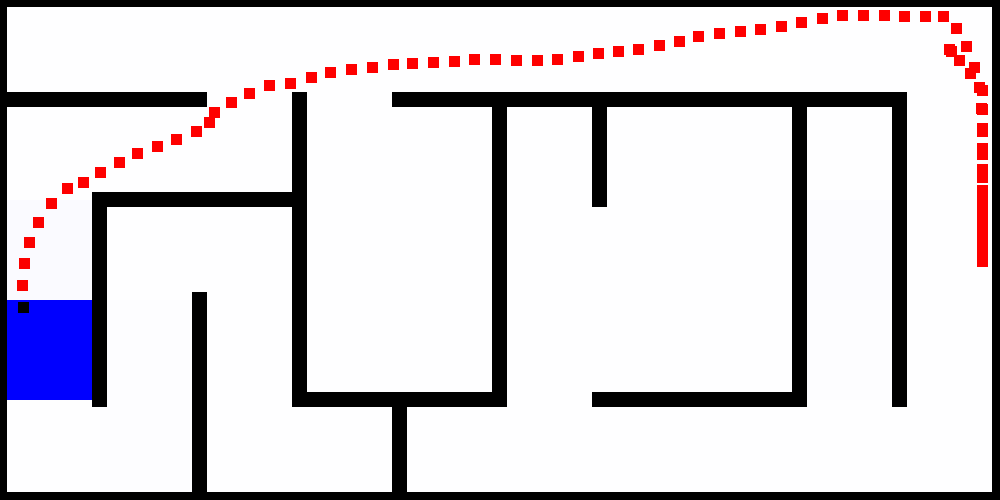}
    \caption{Trajectory of the Nav A3C+$D_2$ agent in the random goal maze 1, overlaid with the position probability predictions predicted by a decoder trained on LSTM hidden activations, taken at 4 steps during an episode. Initial uncertainty gives way to accurate position prediction as the agent navigates.}
    \label{fig:posdecoding}
\end{figure}

An example of position decoding by the Nav A3C+$D_2$ agent is shown in Figure \ref{fig:posdecoding}, where the initial uncertainty in position is improved to near perfect position prediction as more observations are acquired by the agent. We observe that position entropy spikes after a respawn, then decreases once the agent acquires certainty about its location. Additionally, videos of the agent's position decoding are linked in Appendix \ref{sec:videos}.
In these complex mazes, where localization is important for the purpose of reaching the goal, it seems that position accuracy and final score are correlated, as shown in Table \ref{tab:position}. 
A pure feed-forward architecture still achieves 64.3\% accuracy in a static maze with static goal, suggesting that the encoder memorizes the position in the weights 
and that this small maze is solvable by all the agents, with sufficient training time. In Random Goal 1, it is Nav A3C+$D_2$ that achieves the best position decoding performance (85.5\% accuracy), whereas the FF A3C and the LSTM A3C architectures are at approximately 50\%.

\begin{figure}
    \centering
    \includegraphics[width=0.26\textwidth,bb=0 0 640 480]{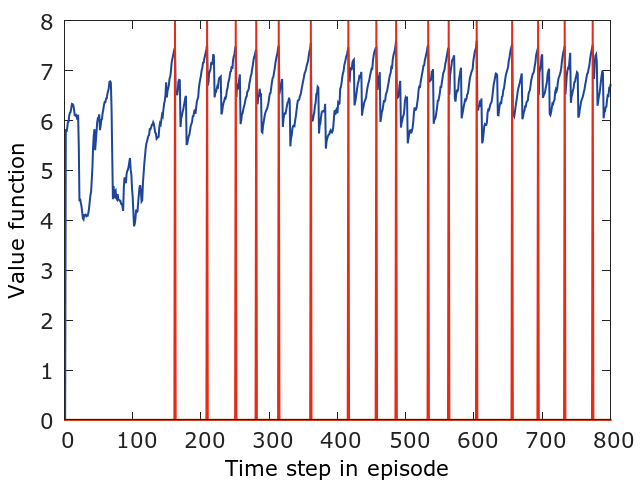} 
    \raisebox{0.25\height}{\includegraphics[width=0.22\textwidth,trim={.5in, 1.2in, .5in, .5in}, clip,bb=0 0 1100 700]{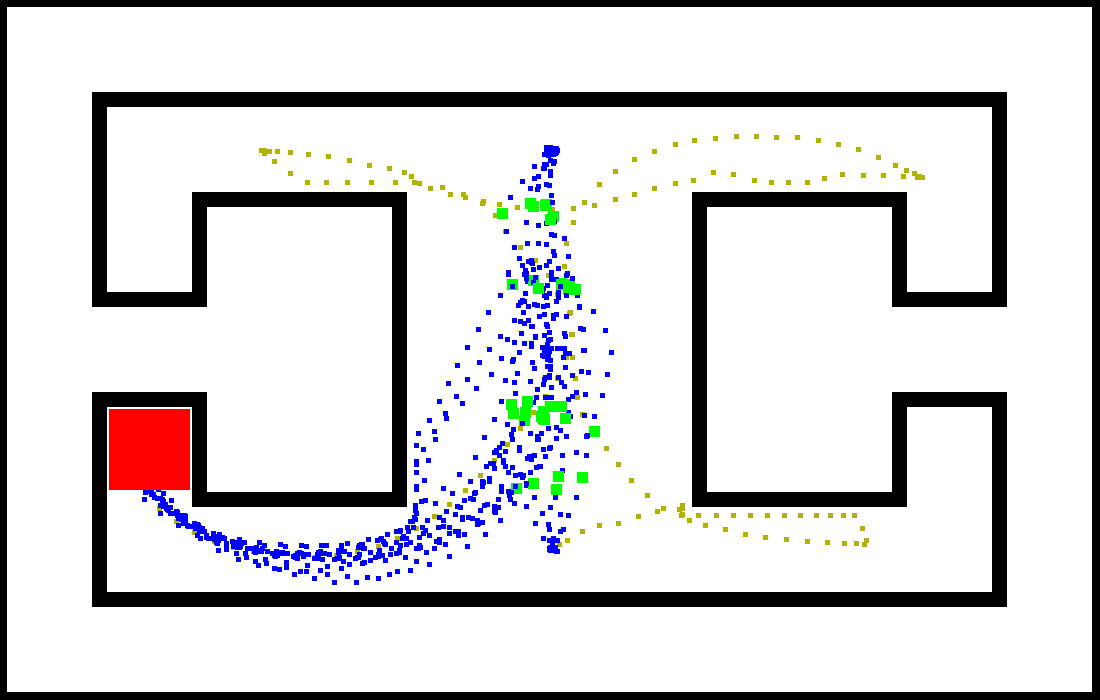}} 
    \includegraphics[width=0.26\textwidth,bb=0 0 640 480]{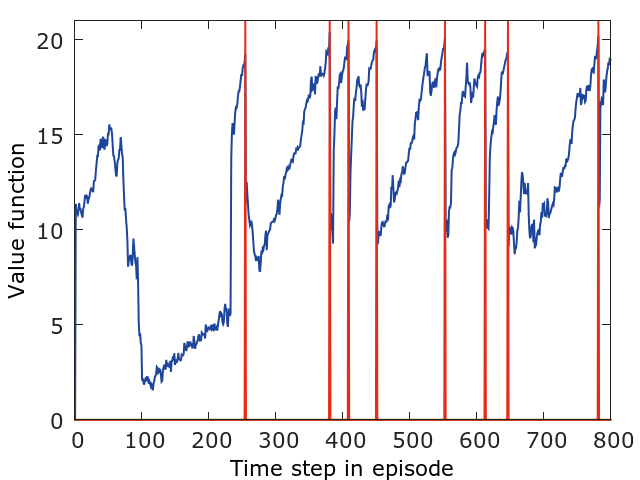} 
    \raisebox{0.5\height}{\includegraphics[width=0.22\textwidth,trim={.5in, 1.2in, .5in, .5in}, clip,bb=0 0 1000 500]{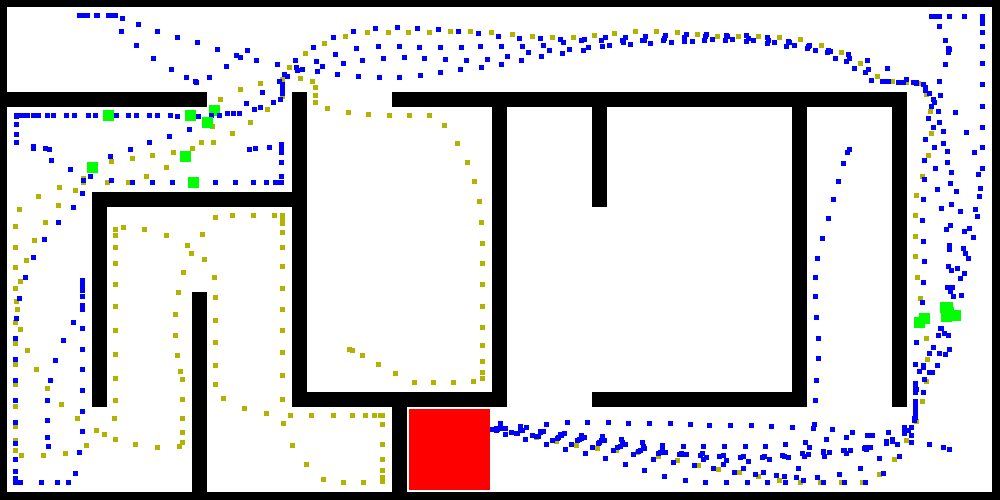}}
    \caption{Trajectories of the Nav A3C*+$D_1L$ agent in the I-maze (left) and of the Nav A3C+$D_2$ random goal maze 1 (right) over the course of one episode. At the beginning of the episode (gray curve on the map), the agent explores the environment until it finds the goal at some unknown location (red box). During subsequent respawns (blue path), the agent consistently returns to the goal. The value function, plotted for each episode, rises as the agent approaches the goal. Goals are plotted as vertical red lines.}
    \label{fig:imazetraj}
\end{figure}

In the I-maze, the opposite branches of the maze are nearly identical, with the exception of very sparse visual cues. We observe that once the goal is first found, the Nav A3C*+$D_1L$ agent is capable of directly returning to the correct branch in order to achieve the maximal score. However, the linear position decoder for this agent
is only 68.5\% accurate, whereas it is 87.8\% in the plain LSTM A3C agent. We hypothesize that the symmetry of the I-maze will induce a symmetric policy that need not be sensitive to the exact position of the agent (see analysis below).

A desired property of navigation agents in our Random Goal tasks is to be able to first find the goal, and reliably return to the goal via an efficient route after subsequent re-spawns. 
The latency column in Table \ref{tab:position} shows that the Nav A3C+$D_2$ agents achieve the lowest latency to goal once the goal has been discovered (the first number shows the time in seconds to find the goal the first time, and the second number is the average time for subsequent finds). Figure \ref{fig:imazetraj} shows clearly how the agent finds the goal, and directly returns to that goal for the rest of the episode. For Random Goal 2, none of the agents achieve lower latency after initial goal acquisition; this is presumably due to the larger, more challenging environment.

\subsection{Stacked LSTM goal analysis}
\begin{figure}
    \centering
    \subfloat[\scriptsize{Agent trajectories for episodes with different goal locations}]{
    \raisebox{.25\height}{\includegraphics[width=.28\textwidth,trim={.5in, 0in, .2in, 0in}, clip,bb=0 0 575 385]{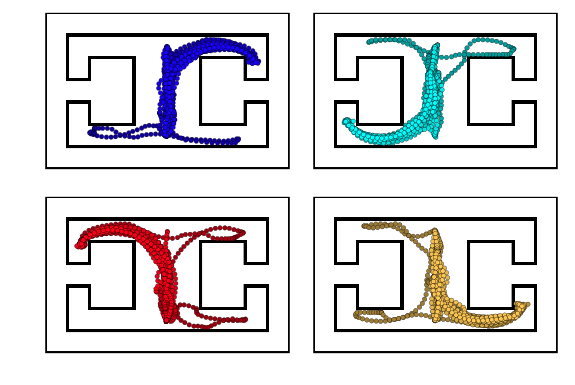}}}
    \hspace{.2in}
    \subfloat[\scriptsize{LSTM activations from A3C agent}]{
    \includegraphics[width=.28\textwidth,trim={.7in, .7in, .7in, .7in}, clip,bb=0 0 556 594]{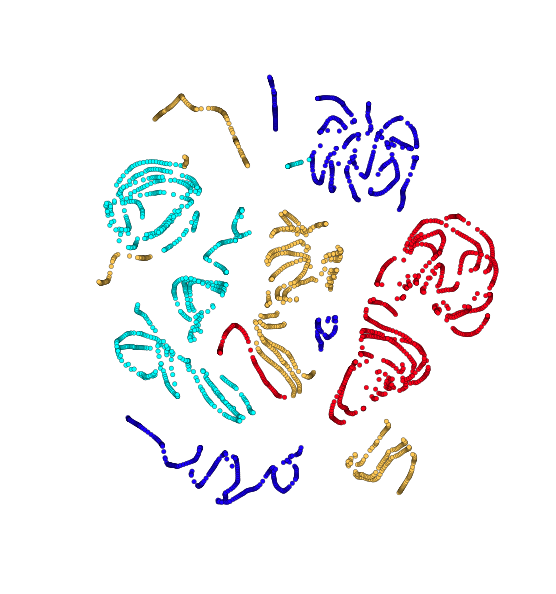}
    }
    \hspace{.2in}
    \subfloat[\scriptsize{LSTM activations from Nav A3C*+$D_1L$ agent}]{
    \includegraphics[width=.28\textwidth,trim={.5in, .7in, .5in, .3in}, clip,bb=0 0 556 594]{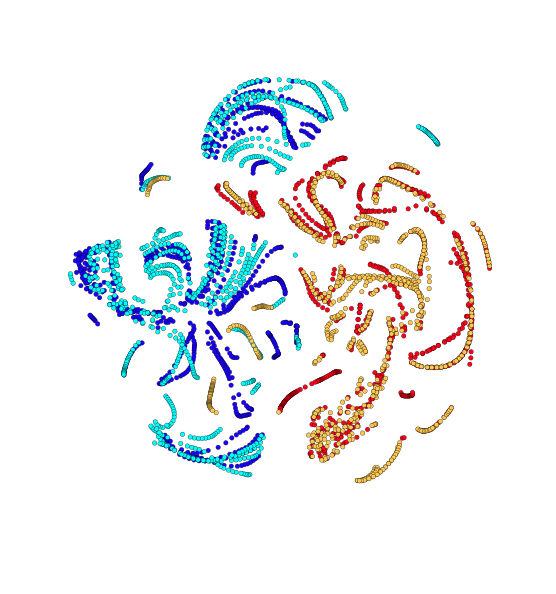}}
    \caption{LSTM cell activations of LSTM A3C and Nav A3C*+$D_1L$ agents from the I-Maze collected over multiple episodes and reduced to 2 dimensions using tSNE, then coloured to represent the goal location. Policy-dictating LSTM of Nav A3C agent shown.}
    \label{fig:tsne}
\end{figure}

Figure \ref{fig:tsne}(a) shows shows the trajectories traversed by an agent for each of the four goal locations. After an initial exploratory phase to find the goal, the agent consistently returns to the goal location. We visualize the agent's policy by applying tSNE dimension reduction \citep{tSNE} to the \emph{cell} activations at each step of the agent for each of the four goal locations. Whilst clusters corresponding to each of the four goal locations are clearly distinct in the LSTM A3C agent, there are 2 main clusters in the Nav A3C agent -- with trajectories to diagonally opposite arms of the maze represented similarly. Given that the action sequence to opposite arms is equivalent (e.g. straight, turn left twice for top left and bottom right goal locations), this suggests that the Nav A3C policy-dictating LSTM maintains an efficient representation of 2 sub-policies (i.e. rather than 4 independent policies) -- with critical information about the currently relevant goal provided by the additional LSTM.


\subsection{Investigating different combinations of auxiliary tasks}
Our results suggest that depth prediction from the policy LSTM yields optimal results.
However, several other auxiliary tasks have been concurrently introduced in \citep{JaderbergDreamingICLR}, and thus we provide a comparison of reward prediction against depth prediction. Following that paper, we implemented two additional agent architectures, one performing reward prediction from the convnet using a replay buffer, called Nav A3C*+$R$, and one combining reward prediction from the convnet and depth prediction from the LSTM (Nav A3C+$RD_2$). Table \ref{tab:optimal} 
suggests that reward prediction (Nav A3C*+$R$) improves upon the plain stacked LSTM architecture (Nav A3C*) but not as much as depth prediction from the policy LSTM (Nav A3C+$D_2$). Combining reward prediction and depth prediction (Nav A3C+$RD_2$) yields comparable results to depth prediction alone (Nav A3C+$D_2$); normalised average AUC values are respectively 0.995 vs. 0.981. Future work will explore other auxiliary tasks.

\begin{table}
\begin{center}
\scriptsize
\begin{tabular}{@{}l|cccccc@{}} 
 \toprule
 & \multicolumn{6}{c}{\textbf{Navigation agent architecture}} \\
 \textbf{Maze} & Nav A3C* & Nav A3C+$D_1$ & Nav A3C+$D_2$ & Nav A3C+$D_1D_2$ & Nav A3C*+$R$ & Nav A3C+$RD_2$ \\ 
 \midrule
 \textbf{I-Maze} & 143.3 & 196.7 & {\bf 203.5} & 197.2 & 128.2 & 191.8 \\
 \textbf{Static 1} & 60.1 & 103.2 & 104.3 & 100.3 & 86.9 & {\bf 105.1} \\
 \textbf{Static 2} & 59.9 & 153.1 & {\bf 157.6} & 151.6 & 100.6 & 155.5 \\
 \textbf{Random Goal 1} & 45.5 & 57.6 & 71.1 & 63.2 & 54.4 & {\bf 72.3} \\
 \textbf{Random Goal 2} & 37.0 & 66.0 & {\bf 82.1} & 75.1 & 68.3 & 80.1 \\
 \bottomrule
\end{tabular}
\end{center}
\caption{Comparison of five navigation agent architectures over five maze configurations with random and static goals, including agents performing reward prediction Nav A3C*+$R$ and Nav A3C+$RD_2$, where reward prediction is implemented following \citep{JaderbergDreamingICLR}. We report the \emph{AUC} (Area under learning curve), averaged over the best 5 hyperparameters.}
\label{tab:optimal}
\end{table}

\section{Conclusion}

We proposed a deep RL method, augmented with memory and auxiliary learning targets, for training agents to navigate within large and visually rich environments that include frequently changing start and goal locations. Our results and analysis highlight the utility of un/self-supervised auxiliary objectives, namely depth prediction and loop closure, in providing richer training signals that bootstrap learning and enhance data efficiency. Further, we examine the behavior of trained agents, their ability to localise, and their network activity dynamics, in order to analyse their navigational abilities.  

Our approach of augmenting deep RL with auxiliary objectives allows end-end learning and may encourage the development of more general navigation strategies.  Notably, our work with auxiliary losses is related to \citep{JaderbergDreamingICLR} which independently looks at data efficiency when exploiting auxiliary losses.
One difference between the two works is that our auxiliary losses are online (for the current frame) and do not rely on any form of replay. Also the explored losses are very different in nature. Finally our focus is on the navigation domain  and understanding if navigation emerges as a bi-product of solving an RL problem, while 
\cite{JaderbergDreamingICLR} is concerned with data efficiency for any RL-task. 

Whilst our best performing agents are relatively successful at navigation, their abilities would be stretched if larger demands were placed on rapid memory (e.g. in procedurally generated mazes), due to the limited capacity of the stacked LSTM in this regard. It will be important in the future to combine visually complex environments with architectures that make use of external memory  \citep{graves2016hybrid, weston2014memory, olton1979hippocampus}  to enhance the navigational abilities of agents. Further, whilst this work has focused on investigating the benefits of auxiliary tasks for developing the ability to navigate through end-to-end deep reinforcement learning, it would be interesting for future work to compare these techniques with SLAM-based approaches. 

ACKNOWLEDGEMENTS

We would like to thank Alexander Pritzel, Thomas Degris and Joseph Modayil for useful discussions, Charles Beattie, Julian Schrittwieser, Marcus Wainwright, and Stig Petersen for environment design and development, and Amir Sadik and Sarah York for expert human game testing.

\bibliography{references}
\bibliographystyle{iclr2017_conference}

\newpage

\appendix
\pagenumbering{arabic}

{\Large{\textbf{Supplementary Material}}}

\section{Videos of trained navigation agents}
\label{sec:videos}

We show the behaviour of Nav A3C*+$D_1L$ agent in 5 videos, corresponding to the 5 navigation environments: I-maze\footnote{Video of the Nav A3C*+$D_1L$ agent on the I-maze: \url{https://youtu.be/PS4iJ7Hk_BU}}, (small) static maze\footnote{Video of the Nav A3C*+$D_1L$ agent on static maze 1: \url{https://youtu.be/-HsjQoIou_c}}, (large) static maze\footnote{Video of the Nav A3C*+$D_1L$ agent on static maze 2: \url{https://youtu.be/kH1AvRAYkbI}}, (small) random goal maze\footnote{Video of the Nav A3C*+$D_1L$ agent on random goal maze 1: \url{https://youtu.be/5IBT2UADJY0}} and (large) random goal maze\footnote{Video of the Nav A3C*+$D_1L$ agent on random goal maze 2: \url{https://youtu.be/e10mXgBG9yo}}. Each video shows a high-resolution video (the actual inputs to the agent are down-sampled to 84$\times$84 RGB images), the value function over time (with fruit reward and goal acquisitions), the layout of the mazes with consecutive trajectories of the agent marked in different colours and the output of the trained position decoder, overlayed on top of the maze layout.

\section{Network Architecture and Training}
\label{sec:app}

\subsection{The online multi-learner algorithm for multi-task learning}

We introduce a class of neural network-based agents that have modular structures and that are trained on multiple tasks, with inputs coming from different modalities (vision, depth, past rewards and past actions). Implementing our agent architecture is simplified by its modular nature. Essentially, we construct multiple networks, one per task, using shared building blocks, and optimise these networks jointly. Some modules, such as the conv-net used for perceiving visual inputs, or the LSTMs used for learning the navigation policy, are shared among multiple tasks, while other modules, such as depth predictor $g_d$ or loop closure predictor $g_l$, are task-specific. The navigation network that outputs the policy and the value function is trained using reinforcement learning, while the depth prediction and loop closure prediction networks are trained using self-supervised learning.

Within each thread of the asynchronous training environment, the agent plays on its own episode of the game environment, and therefore sees observation and reward pairs $\{(s_t, r_t)\}$ and takes actions that are different from those experienced by agents from the other, parallel threads. Within a thread, the multiple tasks (navigation, depth and loop closure prediction) can be trained at their own schedule, and they add gradients to the shared parameter vector as they arrive. Within each thread, we use a flag-based system to subordinate gradient updates to the A3C reinforcement learning procedure.

\subsection{Network and training details}

\begin{figure}
    \centering
    \includegraphics[width=0.5\textwidth]{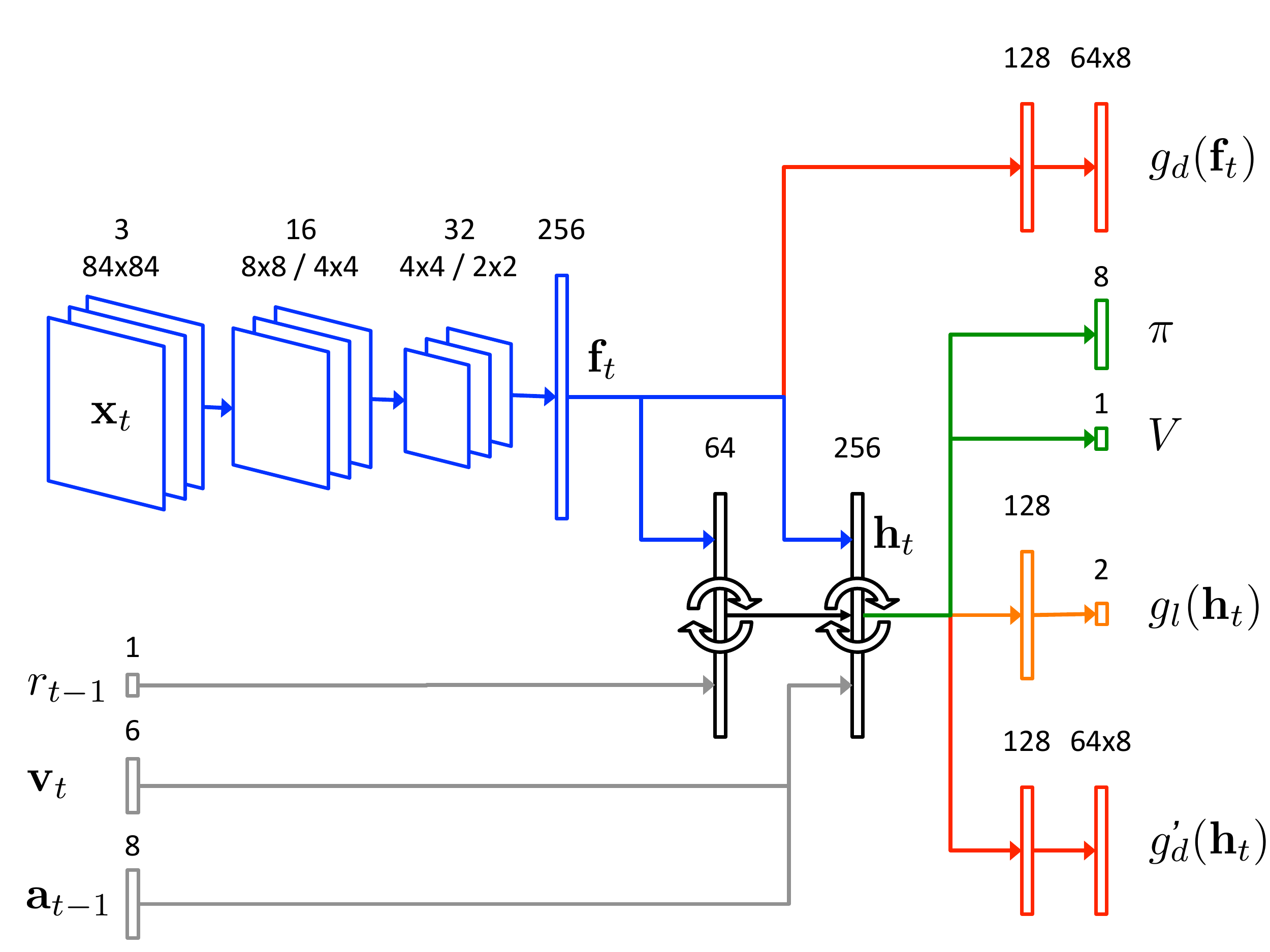}
    \caption{Details of the architecture of the Nav A3C+D+L+Dr agent, taking in RGB visual inputs ${\bf x}_t$, past reward $r_{t-1}$, previous action ${\bf a}_{t-1}$ as well as agent-relative velocity ${\bf v}_t$, and producing policy $\pi$, value function $V$, depth predictions $g_d({\bf f}_t)$ and $g'_d({\bf h}_t)$ as well as loop closure detection $g_l({\bf h}_t)$.}
    \label{fig:archinava3c}
\end{figure}

For all the experiments we use an encoder model with 2 convolutional layers followed by a fully connected layer, or recurrent layer(s), from which we predict the policy and value function. The architecture is similar to the one in \citep{mnih2016a3c}. The convolutional layers are as follows. The first convolutional layer has a kernel of size 8x8 and a stride of 4x4, and 16 feature maps. The second layer has a kernel of size 4x4 and a stride of 2x2, and 32 feature maps. The fully connected layer, in the FF A3C architecture in Figure \ref{fig:archs}a has 256 hidden units (and outputs visual features $f_t$). The LSTM in the LSTM A3C architecture has 256 hidden units (and outputs LSTM hidden activations $h_t$). The LSTMs in Figure \ref{fig:archs}c and \ref{fig:archs}d are fed extra inputs (past reward $r_{t-1}$, previous action ${\bf a}_t$ expressed as a one-hot vector of dimension 8 and {\it agent-relative} lateral and rotational velocity ${\bf v}_t$ encoded by a 6-dimensional vector), which are all concatenated to vector $f_t$. The Nav A3C architectures (Figure \ref{fig:archs}c,d) have a first LSTM with 64 or 128 hiddens and a second LSTM with 256 hiddens. The depth predictor modules $g_d$, $g'_d$ and the loop closure detection module $g_l$ are all single-layer MLPs with 128 hidden units. The depth MLPs are followed by 64 independent 8-dimensional softmax outputs (one per depth pixel). The loop closure MLP is followed by a 2-dimensional softmax output. We illustrate on Figure \ref{fig:archinava3c} the architecture of the Nav A3C+D+L+Dr agent.

Depth is taken as the Z-buffer from the Labyrinth environment (with values between 0 and 255), divided by 255 and taken to power 10 to spread the values in interval $[0, 1]$. We empirically decided to use the following quantization: $\{0, 0.05, 0.175, 0.3, 0.425, 0.55, 0.675, 0.8, 1\}$ to ensure a uniform binning across 8 classes. The previous version of the agent had a single depth prediction MLP $g_d$ for regressing $8\times16=128$ depth pixels from the convnet outputs ${\bf f}_t$.

The parameters of each of the modules point to a subset of a common vector of parameters. We optimise these parameters using an asynchronous version of RMSProp \citep{Tieleman2012}. \citep{Nair2015} was a recent example of asynchronous and parallel gradient updates in deep reinforcement learning; in our case, we focus on the specific Asynchronous Advantage Actor Critic (A3C) reinforcement learning procedure in \citep{mnih2016a3c}.

Learning follows closely the paradigm described in \citep{mnih2016a3c}. We use 16 workers and the same RMSProp algorithm without momentum or centering of the variance. Gradients are computed over non-overlaping chunks of the episode. The score for each point of a training curve is the average over all the episodes the model gets to finish in $5e4$ environment steps.

The whole experiments are run for a maximum of $1e8$ environment step. The agent has an action repeat of 4 as in \citep{mnih2016a3c}, which means that for 4 consecutive steps the agent will use the same action picked at the beginning of the series. For this reason through out the paper we actually report results in terms of agent perceived steps rather than environment steps. That is, the maximal number of agent perceived step that we do for any particular run is $2.5e7$.

In our grid we sample hyper-parameters from categorical distributions:
\begin{itemize}
  \item Learning rate was sampled from $[10^{-4}, 5\cdot 10^{-4}]$.
  \item Strength of the entropy regularization from $[10^{-4}, 10^{-3}]$.
  \item Rewards were not scaled and not clipped in the new set of experiments. In our previous set of experiments, rewards were scaled by a factor from $\{0.3, 0.5\}$ and clipped to 1 prior to back-propagation in the Advantage Actor-Critic algorithm.
  \item Gradients are computed over non-overlaping chunks of 50 or 75 steps of the episode. In our previous set of experiments, we used chunks of 100 steps.
\end{itemize}

The auxiliary tasks, when used, have hyperparameters sampled from:
\begin{itemize}
  \item Coefficient $\beta_d$ of the depth prediction loss from convnet features $\mathcal{L}_d$ sampled from $\{3.33, 10, 33\}$.
  \item Coefficient $\beta_d'$ of the depth prediction loss from LSTM hiddens $\mathcal{L}_{d'}$ sampled from $\{1, 3.33, 10\}$.
  \item Coefficient $\beta_l$ of the loop closure prediction loss $\mathcal{L}_l$ sampled from $\{1, 3.33, 10\}$.
\end{itemize}

Loop closure uses the following thresholds: maximum distance for position similarity $\eta_1=1$ square and minimum distance for removing trivial loop-closures $\eta_2=2$ squares.






\section{Additional results}
\label{sec:additionalResults}

\begin{figure}
    \centering
    \subfloat[\scriptsize{Random Goal maze (small): comparison of reward clipping}]{
    \includegraphics[width=0.48\textwidth,bb=0 0 494 355]{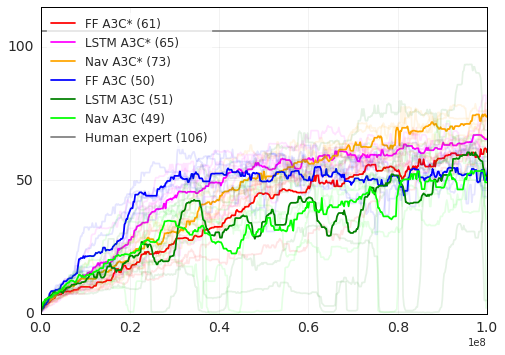}
    }
    \subfloat[\scriptsize{Random Goal maze (small): comparison of depth prediction}]{
    \includegraphics[width=0.48\textwidth,bb=0 0 494 355]{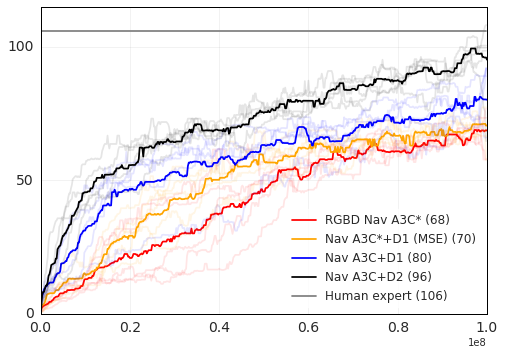}
    }
    \caption{Results are averaged over the top 5 random hyperparameters for each agent-task configuration. Star in the label indicates the use of reward clipping. Please see text for more details. }
    \label{fig:addRewards}
\end{figure}

\subsection{Reward clipping}

Figure~\ref{fig:addRewards} shows additional learning curves. In particular in the left plot we show that the baselines (A3C FF and A3C LSTM) as well as Nav A3C agent without auxiliary losses, perform worse without reward clipping than with reward clipping. It seems that removing reward clipping makes learning unstable in absence of auxiliary tasks. For this particular reason we chose to show the baselines with reward clipping in our main results.

\subsection{Depth prediction as regression or classification tasks}

The right subplot of Figure~\ref{fig:addRewards} compares having depth as an input versus as a target. Note that using RGBD inputs to the Nav A3C agent performs even worse than predicting depth as a regression task, and in general is worse than predicting depth as a classification task. 

\subsection{Non-navigation tasks in 3D maze environments}

\begin{figure}
    \centering
    \subfloat[\scriptsize{Seek-Avoid (learning curves)}]{
        \includegraphics[width=0.45\textwidth,bb=0 0 485 355]{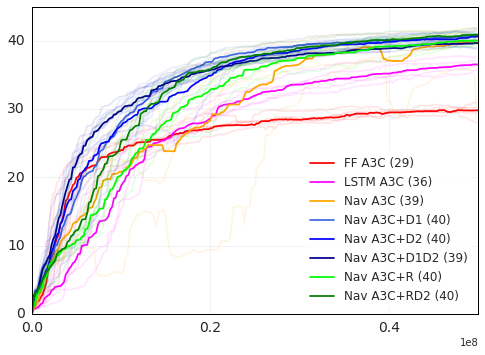}
    }
    \subfloat[\scriptsize{Stairway to Melon (learning curves)}]{
        \includegraphics[width=0.45\textwidth,bb=0 0 485 355]{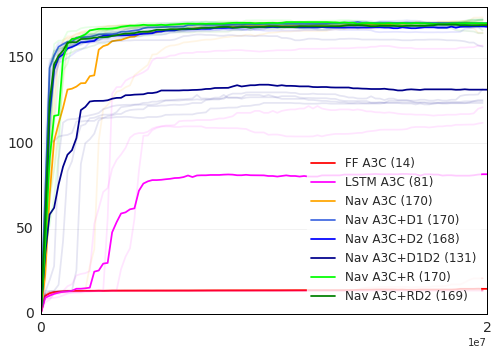}
    } \\
    \subfloat[\scriptsize{Seek-Avoid (layout)}]{
        \includegraphics[width=0.45\textwidth,bb=0 0 919 726]{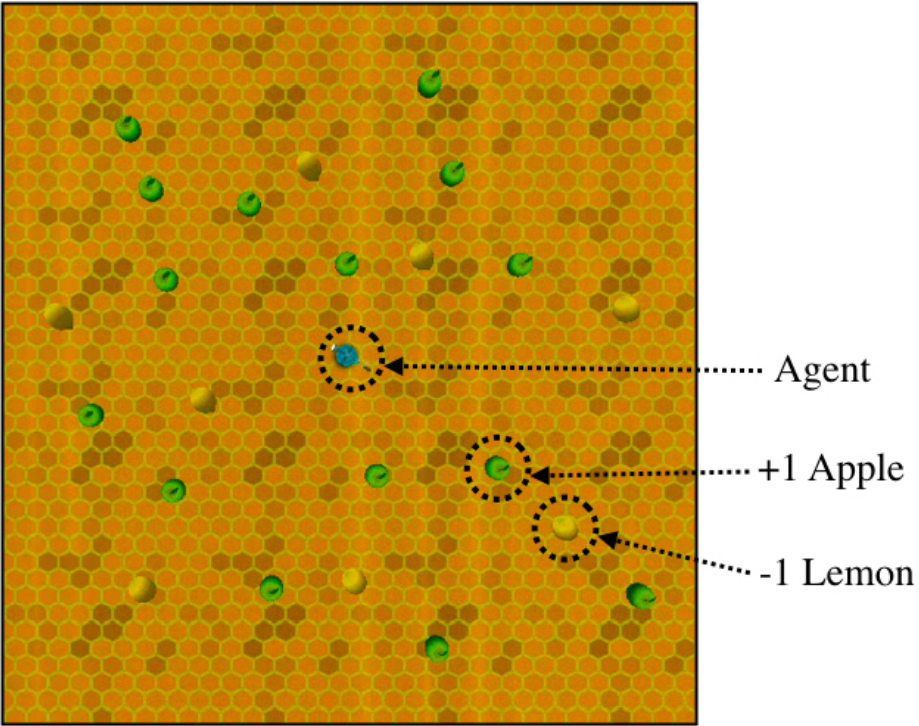}
    }
    \subfloat[\scriptsize{Stairway to Melon (layout)}]{
        \includegraphics[width=0.355\textwidth,bb=0 0 917 923]{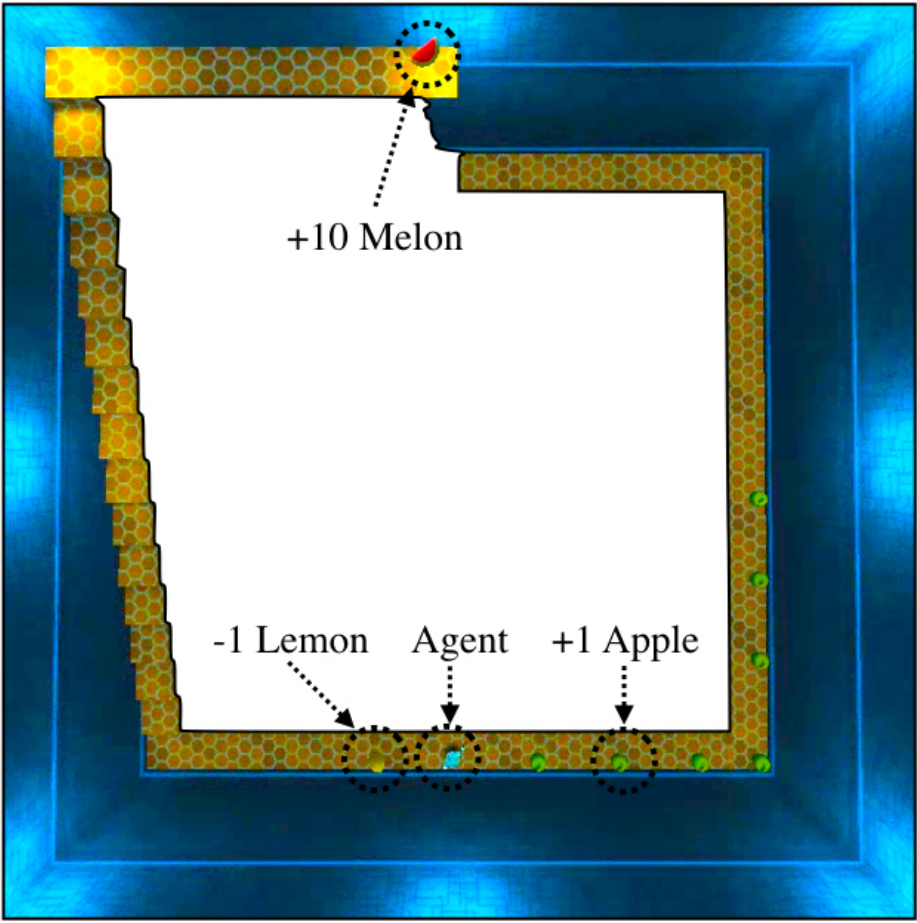}
    }\\
    \caption{Comparison of agent architectures over non-navigation maze configurations, Seek-Avoid Arena and Stairway to Melon, described in details in \citep{Beattie2016}. Image credits for (c) and (d): \citep{JaderbergDreamingICLR}.}
    \label{fig:nonnav}
\end{figure}

We have evaluated the behaviour of the agents introduced in this paper, as well as agents with reward prediction, introduced in \citep{JaderbergDreamingICLR} (Nav A3C*+$R$) and with a combination of reward prediction from the convnet and depth prediction from the policy LSTM (Nav A3C+$RD_2$), on different 3D maze environments with non-navigation specific tasks. In the first environment, Seek-Avoid Arena, there are apples (yielding 1 point) and lemons (yielding -1 point) disposed in an arena, and the agents needs to pick all the apples before respawning; episodes last 20 seconds. The second environment, Stairway to Melon, is a thin square corridor; in one direction, there is a lemon followed by a stairway to a melon (10 points, resets the level) and in the other direction are 7 apples and a dead end, with the melon visible but not reachable. The agent spawns between the lemon and the apples with a random orientation. Both environments have been released in DeepMind Lab \citep{Beattie2016}. These environments do not require navigation skills such as shortest path planning, but a simple reward identification (lemon vs. apple or melon) and persistent exploration. As Figure \ref{fig:nonnav} shows, there is no major difference between auxiliary tasks related to depth prediction or reward prediction. Depth prediction boosts the performance of the agent beyond that of the stacked LSTM architecture, hinting at a more general applicability of depth prediction beyond navigation tasks.


\subsection{Sensitivity towards hyper-parameter sampling}

\begin{figure}
    \centering
    \subfloat[\scriptsize{Static maze (small)}]{
        \includegraphics[width=0.33\textwidth,bb=0 0 505 355]{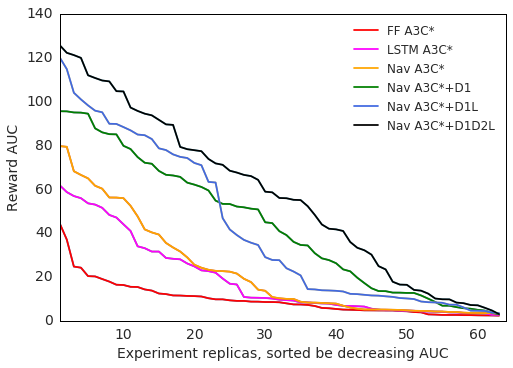}
    }
    \subfloat[\scriptsize{Random Goal maze (large)}]{
        \includegraphics[width=0.33\textwidth,bb=0 0 494 355]{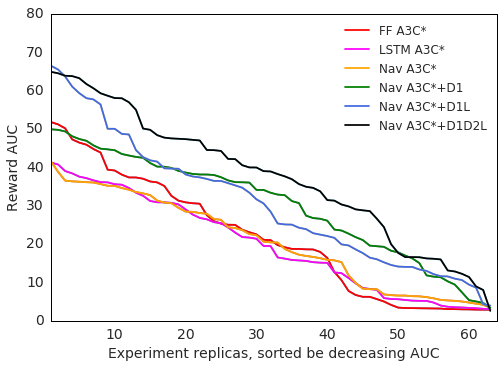}
    }
    \subfloat[\scriptsize{Random Goal I-maze}]{
        \includegraphics[width=0.33\textwidth,bb=0 0 505 355]{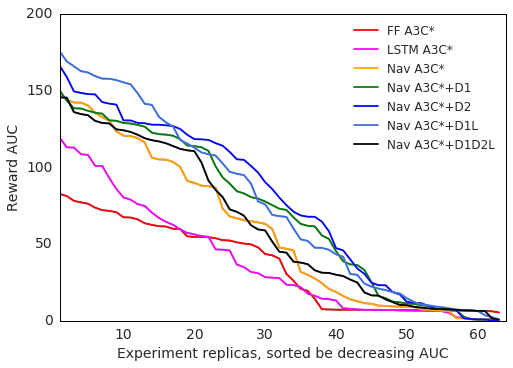}
    }\\
    \caption{Plot of the Area Under the Curve (AUC) of the rewards achieved by the agents, across different experiments and on 3 different tasks: large static maze with fixed goals, large static maze with comparable layout but with dynamic goals, and the I-maze. The reward AUC values are computed for each replica; 64 replicas were run per experiment and the reward AUC values are sorted by decreasing value.}
    \label{fig:aucauc}
\end{figure}

For each of the experiments in this paper, 64 replicas were run with hyperparameters (learning rate, entropy cost) sampled from the same interval. Figure \ref{fig:aucauc} shows that the Nav architectures with auxiliary tasks achieve higher results for a comparatively larger number of replicas, hinting at the fact that auxiliary tasks make learning more robust to the choice of hyperparameters.

\subsection{Asymptotic performance of the agents}

Finally, we compared the asymptotic performance of the agents, both in terms of navigation (final rewards obtained at the end of the episode) and in terms of their representation in the policy LSTM. Rather than visualising the convolutional filters, we quantify the change in representation, with and without auxiliary task, in terms of position decoding, following the approach explained in Section \ref{ref:pos}. Specifically, we compare the baseline agent (LSTM A3C*) to a navigation agent with one auxiliary task (depth prediction), that gets about twice as many gradient updates for the same number of frames seen in the environment: once for the RL task and once for the auxiliary depth prediction task. As Table \ref{tab:asymptotic} shows, the performance of the baseline agent as well as the position decoding accuracy do significantly increase after twice the number of training steps (going from 57 points to 90 points, and from 33.4\% to 66.5\%, but do not reach the performance and position decoding accuracy of the Nav A3C+$D_2$ agent after half the number of training frames. For this reason, we believe that the auxiliary task do more than simply accelerate training.

\begin{table}
\begin{center}
\scriptsize
\begin{tabular}{@{}lc|cc@{}} 
 \toprule
 & & \multicolumn{2}{c}{\textbf{Agent architecture}} \\
 \textbf{Frames} & \textbf{Performance} & LSTM A3C* & Nav A3C+$D_2$ \\ 
 \midrule
 \textbf{120M} & \textbf{Score (mean top 5)} & 57 & {\bf 103} \\
 & \textbf{Position Acc} & 33.4 & {\bf 72.4} \\
 \midrule
 \textbf{240M} & \textbf{Score (mean top 5)} & 90 & {\bf 114} \\
 & \textbf{Position Acc} & 64.1 & {\bf 80.6} \\
 \bottomrule
\end{tabular}
\caption{Asymptotic performance analysis of two agents in the Random Goal 2 maze, comparing training for 120M Labyrinth frames vs. 240M frames.}
\label{tab:asymptotic}
\end{center}
\end{table}

\end{document}